\documentclass{article}
\usepackage{ijcai22}

\usepackage{times}
\usepackage{soul}
\usepackage{url}
\usepackage[hidelinks]{hyperref}
\usepackage[utf8]{inputenc}
\usepackage[small]{caption}
\usepackage{graphicx}
\usepackage{amsmath}
\usepackage{amsthm}
\usepackage{booktabs}
\usepackage{algorithm}
\usepackage{bm}
\usepackage{algorithmic}
\usepackage{amssymb}
\usepackage{multirow}
\usepackage{hyperref}
\hypersetup{colorlinks,
	linkcolor=black,
	anchorcolor=black,
	citecolor=black}

\title{Constrained Adaptive Projection with Pretrained Features for Anomaly Detection}

\author{
Xingtai Gui$^1$
\and
Di Wu$^1$
\and
Shicai Fan$^{1,2}$\\
\affiliations
$^1$School of Automation Engineering, University of Electronic Science and Technology of China(UESTC)\\
$^2$Shenzhen Institute of Advanced Study, UESTC\\
\emails
\{tabgui,wudi123 \}@std.uestc.edu.cn,
shicaifan@uestc.edu.cn
}

\begin{document}

\maketitle

\begin{abstract}
Anomaly detection aims to separate anomalies from normal samples, and the pretrained network is promising for anomaly detection. However, adapting the pretrained features would be confronted with the risk of pattern collapse when finetuning on one-class training data. In this paper, we propose an anomaly detection framework called constrained adaptive projection with pretrained features (CAP). Combined with pretrained features, a simple linear projection head applied on a specific input and its k most similar pretrained normal representations is designed for feature adaptation, and a reformed self-attention is leveraged to mine the inner-relationship among one-class semantic features. A loss function is proposed to avoid potential pattern collapse. Concretely, it considers the similarity between a specific data and its corresponding adaptive normal representation, and incorporates a constraint term slightly aligning pretrained and adaptive spaces. Our method achieves state-of-the-art anomaly detection performance on semantic anomaly detection and sensory anomaly detection benchmarks including 96.5\% AUROC on CIFAR-100 dataset, 97.0\% AUROC on CIFAR-10 dataset and 89.9\% AUROC on MvTec dataset.

\end{abstract}

\section{Introduction}
Anomaly is the pattern that deviates considerably from the concept of normality and detecting such pattern is of key importance in science and industry. Anomaly detection (AD) is a specific task designed to learn a model that accurately detects anomalous test samples \cite{review}. The anomaly may appear due to either covariate shift or semantic shift and these shifts lead to two sub-tasks: sensory AD or semantic AD respectively \cite{review2}.

In standard AD settings, labeled anomalous data are often nonexistent and only normal data are accessible. In such one-class training case, self-supervised methods based on auxiliary tasks, or unsupervised methods like autoencoder are widely used in AD. However, for self-supervised methods, the features trained on auxiliary domains may not generalize well to the target domain and for autoencoder-based methods, their generalization ability could reconstruct the abnormal inputs well and lead to a misjudgment on anomalies.

Due to the restriction of one-class training data, the learned representation is indistinguishable to some extent resulting in the limited performance for anomaly detection. Recently, anomaly detection based on pretrained features has been widely studied. Some works \cite{bergmann2020uninformed,salehi2021multiresolution} only consider leveraging knowledge distillation to transfer the pretrained features of anomaly-free data to the student network, However, they do not adapt the features to the target data set. PANDA \cite{reiss2021panda} proposes a baseline based on pretrained network, and implements the finetune by referring to Deep-SVDD \cite{ruff2018deep}. But such finetune method will cause pattern collapse which means all features will shrink to the center point (See Appendix A for details about pattern collapse). PANDA leverages elastic weight consolidation to restrict the weight change in finetune stage. However, it still needs to further pretrain on an auxiliary task to obtain a Fisher information matrix.

In this work, we introduce a simple yet effective framework considering pretrained feature adaptation which is more suitable for anomaly detection. In order to avoid potential pattern collapse, the framework abandons the traditional optimization with a global parameter such as mean center but adopts a strategy paying attention to normal patterns locally. For a specific input image, its k nearest normal representations in pretrained feature space are traced. Instead of finetuning the layers in pretrained network, we propose a simple adaptive projection head that can play a role in adapting pretrained features properly and map the input and its k nearest normal pretrained features. Self-attention \cite{vaswani2017attention} is reformed to obtain the weights of the projected normal features whose weighted average is regarded as an adaptive normal representation of the input. As a novel anomaly criterion, the similarity between the projected feature of input and the corresponding normal representation is optimized. In addition, based on such framework, we propose a constraint term considering an alignment between the pretrained and adaptive feature spaces to ensure obtaining nontrivial solutions. The joint optimization of the similarity and constraint can avoid pattern collapse and offer promising detection performance.

We summarize our contributions as follows: i) we design an anomaly detection framework constrained adaptive projection with pretrained feature (CAP). Instead of finetuning the large pretrained network, the pretrained features adaptation is with a simple projection head. ii) under the one-class setting, self-attention is reformed to mine inner-relationship between projected k nearest normal features to bring semantic interpretability and enhance the quality of adaptive representation. iii) we propose a novel criterion for anomaly detection and an effective constraint avoiding learning trivial solution. Consequently, without global optimization goal, it can reduce the risk of pattern collapse in optimization.  iv) extensive experiments and visualizations validate the effectiveness of our proposed framework for both sensory AD and semantic AD. The source code of CAP is released at \textit{\textcolor{red}{\url{https://github.com/TabGuigui/CAP}}}.

\section{Related work}
\paragraph{Anomaly Detection} 

Anomaly detection aims to detect any anomalous samples that are deviated from the predefined normality during testing \cite{review2}. Considering traditional anomaly detection with specific machine learning model, kNN \cite{angiulli2002fast} was used to detect outliers. Combined with one-class classification, OC-SVM \cite{scholkopf2001estimating} was proposed to obtain a discriminative hyperplane for anomaly detection. Based on statistical methods, such as PCA \cite{ding2013compressed} and GMM \cite{5}, some suitable models were established to fit the characteristic distribution of normal data. However, it is difficult for such classical methods to work satisfactorily when facing high-dimensional data, such as image anomaly detection tasks.

\paragraph{Deep-learning Based Anomaly Detection} 
Recently, deep learning has pushed the performance of computer vision systems to soaring heights on a broad array of high-level problems. Many deep learning methods have been introduced into anomaly detection community. Ruff et al. proposed Deep-SVDD \cite{ruff2018deep} and  Deep SAD \cite{ruff2019deep} to tackle anomaly detection in unsupervised and semi-supervised way specifically. Gong et al. \cite{gong2019memorizing} added memory module on the basis of autoencoders so as to enlarge the reconstruction errors of abnormal samples for detection. Perera et al. \cite{10} proposed OCGAN method, which would find a potential space to make the reconstructed image of the generator similar to the normal sample. Combined with self-supervised learning, MHRot \cite{hendrycks2019using} was proposed to learn more effective representations leveraging auxiliary tasks.

\paragraph{Transfer-learning Based Anomaly Detection} 
Since anomaly detection is a data-poor task, representations learned on extensive datasets can be leveraged to tackle the limitation. The experiments of Bergman et al. \cite{14} proved that the performance of many self-supervised learning methods applied in image anomaly detection in recent years is far inferior to the simple method based on pretrained strong feature extractor. Salehi et al. \cite{salehi2021multiresolution} designed an anomaly location method utilizing the different behaviors of the clone network distilled by knowledge of pretrained expert network to detect and locate anomalies. Dong et al. \cite{dong2021and} designed a new unsupervised semantic transfer model to explore domain-invariant knowledge between different data distributions. Reiss et al. \cite{reiss2021panda} proposed an approach to adapt pretraining features and suggested some training methods to mitigate pattern collapse.

\section{Our Approach}
\begin{figure*}[h]
	\centering
		\includegraphics[scale=.073]{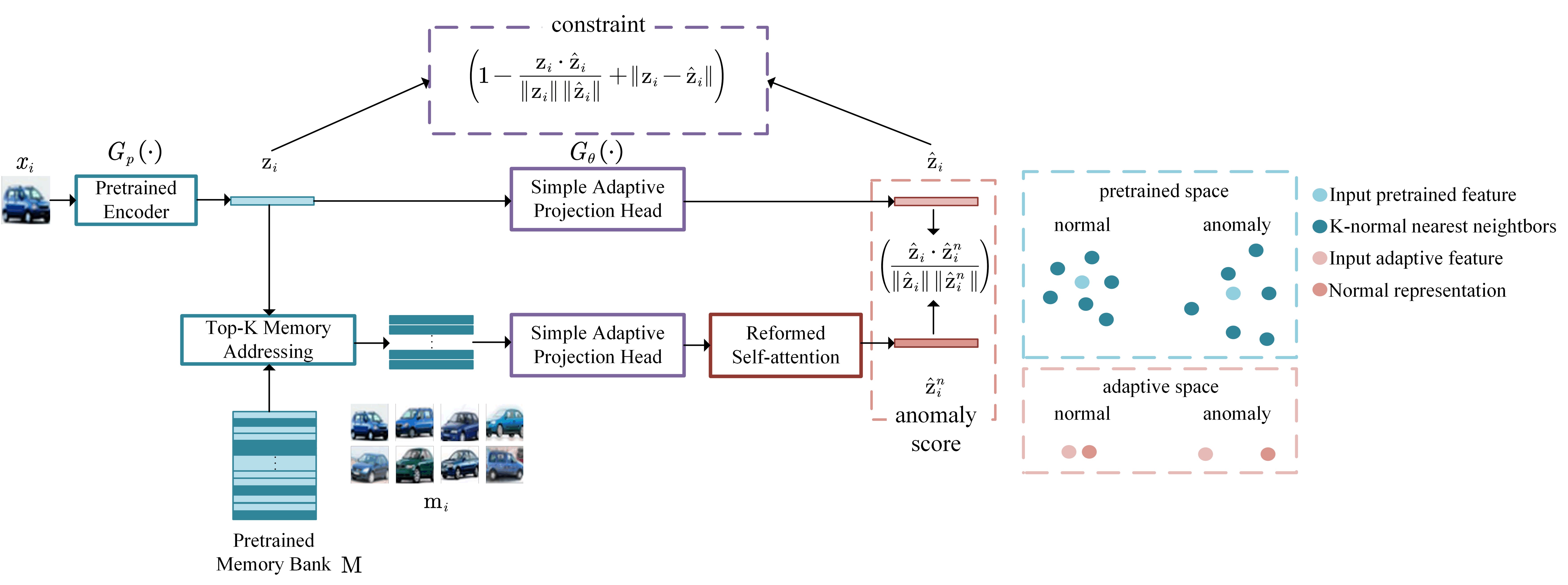}
	\caption{\textbf{Left}: The architecture of CAP contains a pretrained network, a K-normal nearest neighbors module, a simple adaptive projection head and a reformed self-attention module. Here, for a specific input, $\mathbf{z_i}$ is the pretrained feature, $\hat{\mathbf{z}}_i $ is the adaptive feature and $\hat{\mathbf{z}}^n_i $ is its corresponding adaptive normal representation. It should be noted that the two projection heads are weight-shared. \textbf{Right}: A toy example of CAP. In the pretrained space, the dots with light color are the input, and the deep ones are the normal training data. In the adaptive space, the distance between the anomaly and its adaptive normal representation is large.}
	\label{fig:framework}
\end{figure*}
CAP designs a simple adaptive projection head and reforms self-attention for feature adaptation. A novel loss function focusing on data and its adaptive normal representation, with an effective constraint term is proposed. The whole framework is illustrated in Figure \ref{fig:framework} which includes a K-normal nearest neighbors module, an adaptive projection head module and a reformed self-attention module.

\subsection{K-normal Nearest Neighbors}

For a specific input, CAP firstly traces its k most similar normal representations in pretrained space. In the specific implementation, instead of exhaustively computing pretrained representations every time, a memory bank is used to store the pretrained features of all normal training samples. The memory items are obtained through a pretrained large network $G_P$ which is regarded as an encoder. A memory addressing module is leveraged to find the k-normal nearest neighbors.

\paragraph{Pretrained Memory Bank}
The memory bank $ \mathbf{M} \in \mathbb{R}^{N \times D} $ is a real-valued matrix containing $ N $ items of fixed dimension $D$ and $\mathbf{m}_i$ represents a memory item. The memory item $\mathbf{m}_i$ is obtained from the encoder $G_P$, $\mathbf{m}_i = G_P(x^{train}_i)$ where $x^{train}_i$ is a training normal image.

\paragraph{Top-K Memory Addressing}
For a specific image $x_i$, we first compute its pretrained encode $\mathbf{z_i} = G_p(x_i)$, and then compare it against all the pretrained memory items. With a cosine similarity $s_{j,i}= \frac{\mathbf{m}_j\cdot\mathbf{z}_i}{\left \| \mathbf{m}_j \right \| \left \| \mathbf{z}_i \right \| }$, the k-normal nearest neighbors in pretrained space are recorded in a subset $ {\mathbf{M}}_i \in \mathbb{R}^{K \times D}$ and would then be used to construct a normal representation. The similarity ranking of memory items corresponding to $\mathbf{z_i}$ is $ \left[ \mathbf{m}_{i1}, \mathbf{m}_{i2},\mathbf{m}_{i3},...,\mathbf{m}_{in} \right]  $ (from the most to the least). It should be noted that in the training phase, $ \mathbf{z}_i = \mathbf{m}_{i1} $, thus the appropriate K-normal nearest neightbors are  $ {\mathbf{M}}_i = \left[ \mathbf{m}_{i2}, \mathbf{m}_{i3}, ..., \mathbf{m}_{i(k+1)} \right]$ while in the test phase, $ {\mathbf{M}}_i = \left[ \mathbf{m}_{i1}, \mathbf{m}_{i2}, ..., \mathbf{m}_{ik} \right]$.

\subsection{Simple Adaptive Projection Head}
Instead of finetuning some layers of the pretrained large network, we add an adaptive projection head $G_{\bm{\theta}}$ to obtain adpative representations. \cite{chen2020simple,tolstikhin2021mlp} proposed that a function including linear transformation and nonlinear activation function can effectively improve the representation ability. In CAP, a simple adaptive projection head merely including linear transformation is leveraged to finetune $\mathbf{z}_i$ and $ {\mathbf{M}}_i $ simultaneously and get their corresponding mapping
\begin{equation}
    G_{\bm{\theta}}(\mathbf{z};W_{\bm{\theta}}) = W_{\bm{\theta}}\mathbf{z} \quad  W_{\bm{\theta}}\in \mathbb{R}^{D \times D}
\end{equation}

\begin{equation}
    \hat{\mathbf{z}}_i = G_{\bm{\theta}}(\mathbf{z}_i;W_{\bm{\theta}}) \quad
    \hat{\mathbf{M}}_i = G_{\bm{\theta}}(\mathbf{M}_i;W_{\bm{\theta}})
\end{equation}

We have found and proved experimentally that in anomaly detection task, a simple one-layer mapping without activation function is more effective than multi-layer mapping. Moreover, for better adapting to target data, the structure of adaptive projection is designed as a square matrix. These contents will be described in detail later.

\subsection{Reformed Self-attention}
For K-normal nearest neighbors, their feature-level mixup can be used as a better normal representation. The general way is with mean value or considering the cosine similarity with a query. However, the relationship among these items is not considered. In CAP, we believe that the nearest k adaptive representations in $ \hat{\mathbf{M}}_i $ can learn the weight considering their inner-relationship. \cite{vaswani2017attention} proposed self-attention to learn the relationship in sequences paralleled and \cite{su2019semantic} applied a batch-attention module to capture the discriminative information from similar objects. We consider that the subset $ \hat{\mathbf{M}}_i $ contains the information that can infer $ \hat{\mathbf{z}}_i $, naturally, we reform the self-attention mechanism to adapt to the one-class scenario. 

\begin{equation}
    A(\hat{\mathbf{M}}_i;W_Q,W_K)={\rm Softmax}(\frac{(\hat{\mathbf{M}}_iW_Q)(\hat{\mathbf{M}}_iW_K)^T}{\sqrt{D}})\hat{\mathbf{M}}_i
\end{equation}
$W_Q$, $W_K \in \mathbb{R}^{D \times D}$ are linear transforms and add sufficient expressive power for normal features in adaptive space. In addition to such architecture, a residual connection and average operation are inserted and the output of such module is
\begin{equation}
    \hat{\mathbf{z}}^n_i = {\rm Mean}(\hat{\mathbf{M}}_i + A(\hat{\mathbf{M}}_i;W_Q,W_K))
\end{equation}
$\hat{\mathbf{z}}^n_i$ is the weighted average of subset $\hat{\mathbf{M}_i}$ whose weight is learned considering the inner-relationship. Compared with the basic self-attention structure, we remove $W_V$ since the data space should keep unchanged. Moreover, the layer normalization is removed because its anti-overfitting property is unfavorable to anomaly detection. Since $\hat{\mathbf{z}}^n_i$ is calculated with transformed normal subset, it can be regarded as an adaptive normal representation.

\subsection{CAP Loss Function}


\begin{table*}[h]
\caption{Anomaly detection performance (Average AUROC \%).}
\label{tab:overall}
\vspace{-0.3cm}
\begin{center}
\setlength{\tabcolsep}{4mm}{
\begin{tabular}{c|c|c|c|c}
\hline
Method       & CIFAR-10 & CIFAR-100 & FMNIST & MvTec \\ \hline
Deep-SVDD\cite{ruff2018deep}    &  64.8       & 67.0         &  84.8      & 77.9      \\
MHRot\cite{hendrycks2019using}        &  90.1         & 80.1         &  93.2      & 65.5      \\
Distillation\cite{salehi2021multiresolution} &   87.2      &   -       & 94.5       & 87.7       \\
DN2\cite{reiss2021panda}        & 92.5        &  94.1        &    94.5    &   86.5    \\
PANDA\cite{reiss2021panda}        & 96.2        &  94.1        &    \textbf{95.6}    &   86.5    \\ \hline
CAP(no adaptation)         & 96.6        & 96.0         & 95.1        &  89.4     \\
CAP         & \textbf{97.0}        &    \textbf{96.5}      & 95.5       &    \textbf{89.9}   \\ \hline

\end{tabular}}
\end{center}
\end{table*}

We make a hypothesis that the nearest k items in the pretrained space are still the nearest ones in the adaptive space. Thus, we look upon $\hat{\mathbf{z}}^n_i$ as the normal representation corresponding to $\hat{\mathbf{z}}_i $.  Naturally, like Figure \ref{fig:framework} right, for a normal sample, $ \hat{\mathbf{z}}_i $ would be similar to $\hat{\mathbf{z}}^n_i$ while for an anomaly, their dissimilarity is large. In the training phase, since all accessible samples are normal, we take the similarity between $ \hat{\mathbf{z}}_i $ and $\hat{\mathbf{z}}^n_i$ as the optimization target. 

\begin{equation}
    L_S = \frac{1}{N}\sum_{i=1}^N (1- \frac{\hat{\mathbf{z}}_i\cdot\hat{\mathbf{z}}^n_i}{\left \| \hat{\mathbf{z}}_i \right \| \left \| \hat{\mathbf{z}}^n_i \right \| })
\end{equation}

Unlike the loss function mentioned in Deep-SVDD and PANDA which reduces the distance between training data and a global target, such loss function optimizes the similarity between a specific input and its k-normal nearest neighbors in the adaptive space. Therefore, global pattern collapse can be avoided to some extent. Considering $L_S$ only, CAP may tend to learn a trivial solution. However, based on such framework, a simple and practicable constraint is proposed to effectively avoid such potential risks. Here we theoretically demonstrate the potential trivial solution and the feasible method.

\paragraph{Proposition} \textit{Let $G_{\bm{\theta}}$ be the all-zero network weights. For this choice of parameters, the adaptive projection head maps any input to the same output, i.e. $G_{\bm{\theta}}(\mathbf{z}_i)= \mathbf{0}$}.

Since $G_{\bm{\theta}}$ is a simple linear projection, $G_{\bm{\theta}} = \mathbf{0}$ is a potential solution. As the output of the all-zero-weights projection is zero for every input, $L_S$ will become meaningless. However, $L_S$ would optimize $G_{\bm{\theta}}$ into a sparse matrix tending to zero which greatly reduces the representation capacity of model, since with such solution, the loss error tends to be minimum.

To avoid such a trivial solution, based on such framework, the adaptation can be constrained by the pretrained space. Specifically, we consider an alignment between pretrained and adaptive spaces. We design $G_{\bm{\theta}}$ as a square matrix and propose a constraint term

\begin{equation}
    \Omega = \frac{1}{N}\sum_{i=1}^N(1- \frac{{\mathbf{z}}_i\cdot\hat{\mathbf{z}}_i}{\left \| {\mathbf{z}}_i \right \| \left \| \hat{\mathbf{z}}_i \right \| } +  \left \| {\mathbf{z}}_i-\hat{\mathbf{z}}_i \right \|_2^2)
\end{equation}

Where we consider the cosine similarity and Euclidean distance of $\mathbf{z}$ and $\hat{\mathbf{z}}_i$  simultaneously. This encourages $\hat{\mathbf{z}}_i$ to be not only close to the $\mathbf{z}$ in terms of Euclidean distance but also be in the same direction. Such constraint, on the one hand, prevents the solution from approaching zero matrix, on the other hand, retains the pretrained information to a certain extent. Combining the aforementioned loss and constraint, $L_{total}$ is formulated as a weighted sum of $L_S$ and $\Omega$

\begin{equation}
    L_{total} = L_s + \lambda\Omega
\end{equation}

For a given test image, we can naturally define an anomaly score by the similarity between its adaptive mapping $\hat{\mathbf{z}}_i $ and adaptive normal representation $\hat{\mathbf{z}}^n_i$

\begin{equation}
    s(\mathbf{x}_i) = 1- \frac{\hat{\mathbf{z}}_i\cdot\hat{\mathbf{z}}^n_i}{\left \| \hat{\mathbf{z}}_i \right \| \left \| \hat{\mathbf{z}}^n_i \right \| }
\end{equation}

\section{Experiments}

In this section, we experimentally evaluate the performance of CAP and use AUROC as an evaluation metric. Four datasets including CIFAR-10, CIFAR-100, FMNIST and MvTec are used as the benchmarks. ResNet152 and WideResNet50 pretrained on ImageNet is leveraged as the pretrained network which is the same as PANDA\cite{reiss2021panda} and the output of its fourth block is regarded as the pretrained feature. Detailed experiment configurations are shown in Appendix B.

\subsection{Comparison with State-of-the-art}

We present anomaly detection performance of CAP compared to the state-of-the-art deep learning based methods: Deep-SVDD \cite{ruff2018deep}, MHRot \cite{hendrycks2019using}, Distillation\cite{salehi2021multiresolution}, DN2 and PANDA\cite{reiss2021panda}. Deep-SVDD is the widely accepted deep-learning based method, MHRot is a high-performance self-supervised method, Distillation, DN2 and PANDA are pretrain based methods. All the results that are available in the original papers are copied exactly. To verify the effectiveness of CAP, we additionally compare the results without adaptation whose anomaly score is the similarity between feature of a specific input and mean value of its K-normal nearest features in pretrained space. 

The performances in different datasets are reported in Table \ref{tab:overall}. We can make the following observations from this table: i) Proposed anomaly detection criterion achieves a significant improvement. Specifically, without adaptation, the anomaly detection performance is better than DN2 which is also a baseline pretrained based method without adaptation. ii) Feature adaptation of a CAP improves the performance. For all datasets, the detection results are about 0.5\% higher than method without adaptation. iii) CAP prominently outperforms the state-of-the-art pretrained based method PANDA (achieving around 0.8\%, 2.4\% and 3.4\% higher on the CIFAR-10, CIFAR-100 and MvTec datasets) while for FMNIST, CAP performs basically the same. Therefore, it demonstrates the effectiveness of the proposed adaptation strategy and anomaly criterion both on semantic AD and sensory AD.

\subsection{Design of Adaptive Projection Head}

For structural design of the simple adaptive projection head, we refer to \cite{chen2020simple}. Concretely, the layer number and nonlinear function would greatly affect the performance. We compare performance with different structure on CIFAR-10, CIFAR-100 and MvTec datasets.

The performances of different head designs are reported in Table \ref{tab:structure}. The result shows that for anomaly detection, the simpler the structure of projection is, the more effective the adaptation would be. In the traditional deep learning tasks, deep model has a certain generalization capability, while for anomaly detection, generalization is disadvantageous where we hope feature adaptation only works in normal data. Thus, in all experiments, the one-layer linear mapping is leveraged as the adaptation function.


\begin{table}[t]
\caption{Anomaly detection performance with different CAP structure design.}
\label{tab:structure}
\setlength{\tabcolsep}{0.75mm}{
\begin{tabular}{c|ccc|c|c}
\hline
Dataset                  & L+ReLU+L & L+ReLU & L & attention & AUROC \\ \hline
\multirow{4}{*}{CIFAR-10} & \checkmark   &        &  & \checkmark&     96.4    \\
                         &          &  \checkmark      & & \checkmark &   96.8      \\
                         &          &        &  \checkmark & \checkmark &   \textbf{97.0}        \\
                         &          &        &  \checkmark &  &    96.8      \\ \hline
\multirow{4}{*}{CIFAR-100} & \checkmark   &        &  & \checkmark&     95.2    \\
&          &  \checkmark      & & \checkmark &   96.3      \\
&          &        &  \checkmark & \checkmark &   \textbf{96.5}        \\
&          &        &  \checkmark &  &    96.2      \\ \hline                         
\multirow{4}{*}{MvTec}   & \checkmark         &        & & \checkmark & 89.5   \\
                         &          &   \checkmark     &  & \checkmark& 89.7         \\
                         &          &        &  \checkmark & \checkmark&    \textbf{89.9}       \\
                         &          &        & \checkmark  & & 89.8      \\ \hline
\end{tabular}}
\end{table}

\subsection{Effect of Reformed Self-attention}
\label{sec:self-attention}
\begin{figure}[t]
	\centering
		\includegraphics[scale=.25]{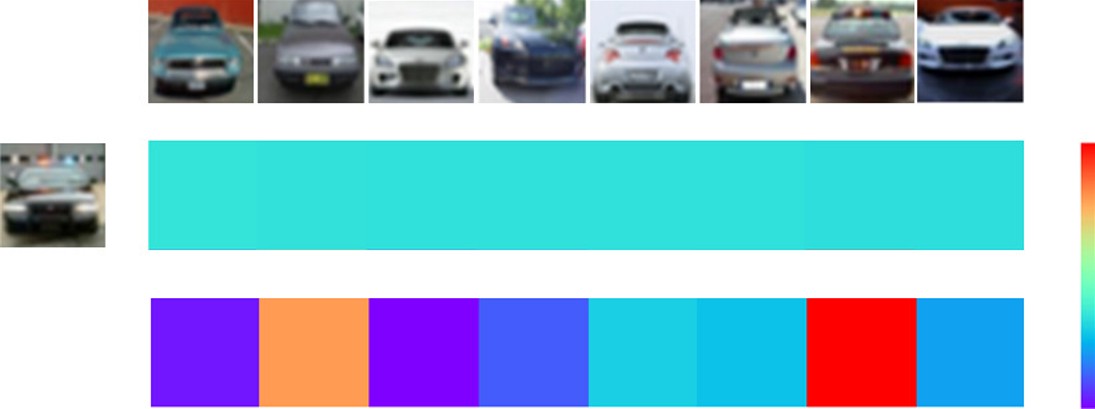}
	\caption{The visualization of effect of reformed self-attention. The left image is a specific input. The middle ones: \textbf{top} the K-normal nearest images in the pretrained space. \textbf{middle} the weight without reformed self-attention. \textbf{bottom} the weight with reformed self-attention.}
	\label{fig:att}
\end{figure}

Self-attention mechanism aims to make the normal representation taking inner-relationship into account. We visualize the effect of reformed self-attention module in Figure \ref{fig:att} where a specific input image, its K-normal nearest images in pretrained space and the weights with or without self-attention are displayed. Moreover, performance with reformed self-attention or not are shown in Table \ref{tab:structure}.

Traditionally, the weights of different items are often obtained by calculating the similarity with the input. We believe the K-normal nearest neighbors contain inner-relationship in one-class setting and the input feature of a specific image should be inferred through such inner-relationship. In Figure \ref{fig:att}, the first weight bar illustrates that weights without self-attention are approximately same while the second bar shows the weights have obvious discrimination. Moreover, the visualization result implies that CAP have considered semantic inner-information. Specifically, The items with larger weight are semantically closest to the input (black cars), while the items with smaller weight are different (cars with other colors). Appendix B provides more case studies.

Anomaly detection performance shown in Table \ref{tab:structure} demonstrates that reformed self-attention module slightly improved adaptation effect. However, such module visually brings more interpretability. It can be concluded that the considering the inner-relationship is reasonable and this module is effective.

\begin{table}[t]
	\centering
	\caption{Anomaly detection performance with different K-normal nearest neighbors (Average AUROC \%).}
	\label{tab:k}
	\setlength{\tabcolsep}{2.5mm}{
		\begin{tabular}{c|cccccc}
			\hline
			K       & 1     & 4     & 8     & 16    & 32    & 64    \\ \hline
			CIFAR-10 & 96.6 & 96.7 & 96.8 & 96.9 & \textbf{97.0} & 97.0 \\
			CIFAR-100 & 96.0 & 96.1 & 96.3 & 96.4 & \textbf{96.5} & 96.4 \\
			MvTec   & 89.5 & \textbf{89.9} & 89.3 & 88.8 & 88.1 & 86.7 \\ \hline
	\end{tabular}}
\end{table}
\subsection{Comparison with Different k Number}

K-normal nearest neighbors is an important module in CAP, in which the number of nearest normal pretrained features should be analyzed. We show performance with different k on CIFAR-10, CIFAR-100 and MvTec in Table \ref{tab:k}.

It can be seen that the performance will be significantly improved when considering multiple pretrained features. For CIFAR-10 and CIFAR-100, a larger k is preferred while for MvTec, k = 4 has obvious advantages. The reason for that is the number of normal training data in MvTec is small where the normal representation of all inputs will be similar if k is large, and resulting in a decrease in the specificity of the adaptive normal representation for each input. There are enough training normal data in CIFAR-10 and CIFAR-100, thus naturally, a larger k promises better performance.

\subsection{Benefit of Constraint Term}
\label{sec:constraint}
We have mentioned a potential trivial solution and proposed a constraint term to avoid such situation. To verify the necessity of the constraint term $\Omega$, we compare the performance with different weights $\lambda$ on CIFAR-10 which is shown in Table \ref{tab:cons}.

\begin{table}[t]
\centering
\caption{Anomaly detection performance on CIFAR-10 with different weights $\lambda$ of constraint term.}
\label{tab:cons}
\setlength{\tabcolsep}{1.7mm}{
\begin{tabular}{c|cccccc|c}
\hline
$\lambda$  & 0    & 0.1  & 1    & 2    & 10 & 100 & no ada \\ \hline
AUROC & 95.1 & 95.6 & 96.9 & 97.0 & 97.0 & 97.0 & 96.6   \\ \hline
\end{tabular}}
\end{table}

\begin{figure}[!t]
	\centering
	\includegraphics[scale=.13]{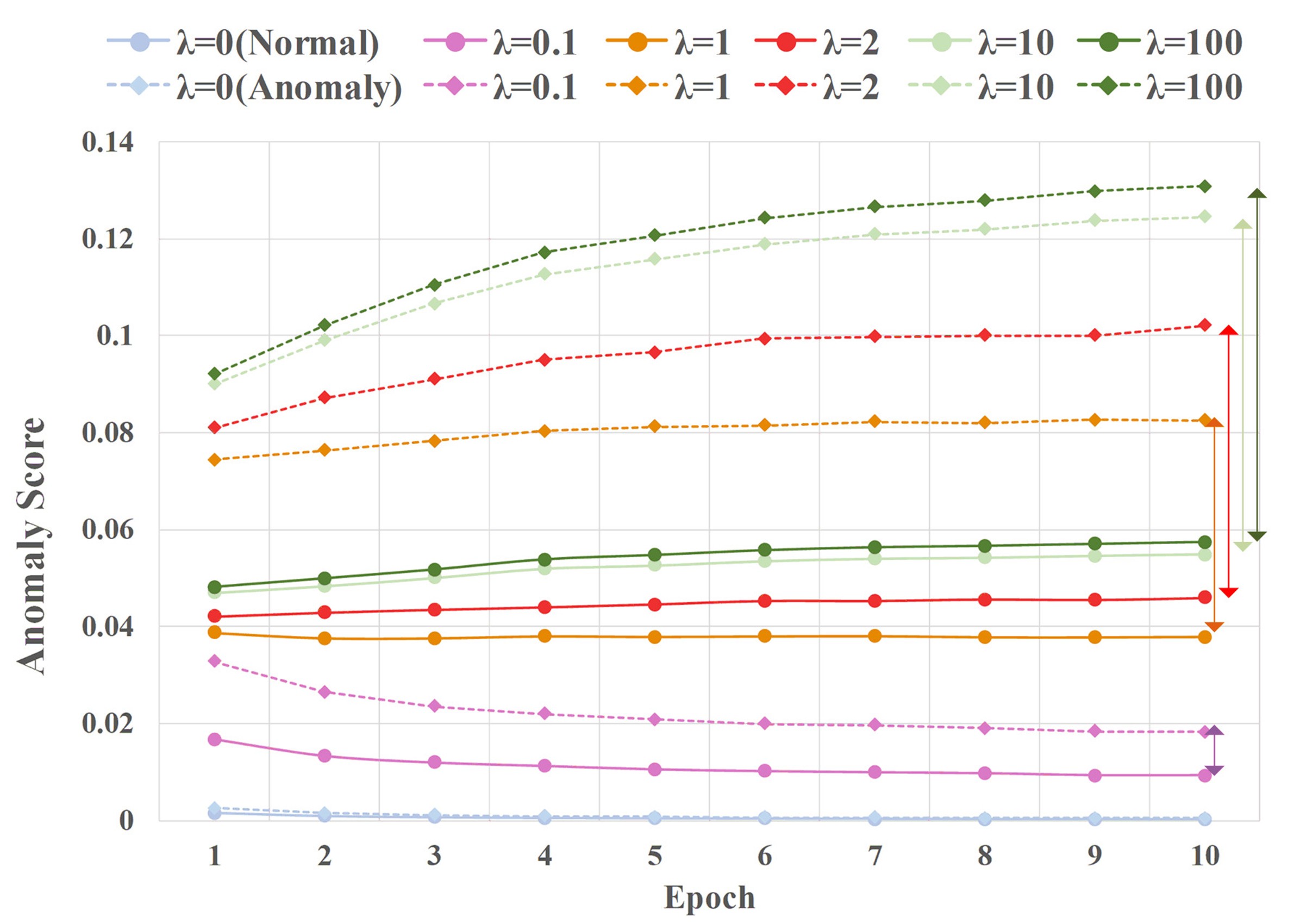}
	\caption{Anomaly score of normal(solid line) and anomaly(dashed line) in test set when training on CIFAR-10 class 0 with different $\lambda$.}
	\label{fig:cons1}
\end{figure}

\begin{figure*}[t]
	\centering
	\includegraphics[scale=.091]{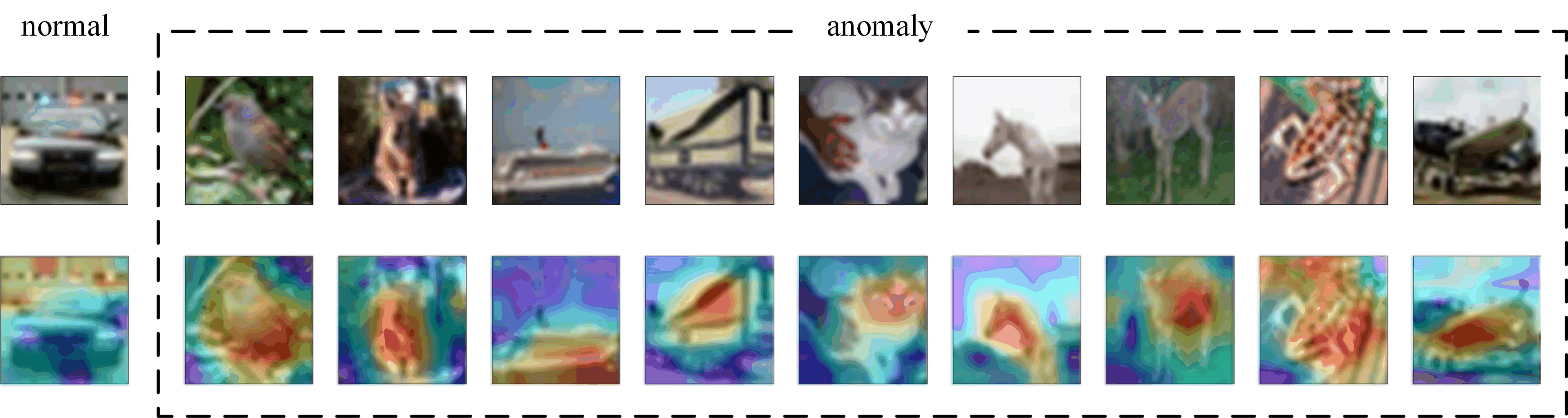}
	\caption{Visualization of CIFAR-10 (Semantic AD). The inferred anomaly region of normal image is in the inessential area and that of anomaly image is on the semantic target.}
	\label{fig:vis1}
\end{figure*}

When $\lambda = 0$, the performance has decreased greatly which shows the potential risk indeed exists. when $\lambda$ is relatively small, the performance has slightly improved, however, it is still inferior to performance without adaptation. While for larger $\lambda$, the model can adapt well, and we find that when the constraint weight reaches a certain value, the result tends to be stable.

To show the benefit of constraint more intuitively, we show detailed anomaly score of normal and anomaly when training on class 0 in Figure \ref{fig:cons1}. For $\lambda = 0$, the anomaly score of all samples are basically equal to $0$ which implies the similarity between $\hat{\mathbf{z}}_i$ and $\hat{\mathbf{z}}^n_i$ is equal to $1$ and the model would indeed tend to learn a trivial solution when there is no constraint. For $\lambda = 0.1 $, there is a gap gradually being small between normal and anomaly which implies an unsatisfactory adaptation. For $\lambda = 1 $ and $\lambda = 2 $, it is shown that the gap of anomaly score between normal and anomaly is gradually growing and a gratifying adaptation appears. The anomaly score of normal is basically unchanged while the anomaly score of anomaly has increased. For larger $\lambda$ like 10 and 100, even though the gap is distinct, the anomaly score of normal rises obviously signifying over-fitting happens on normal sample (See Appendix B for more case studies). These results prove the necessity of constraint term and a slight alignment is promising.

\subsection{Visualization}

\begin{figure}[h]
\centering
	\includegraphics[scale=.105]{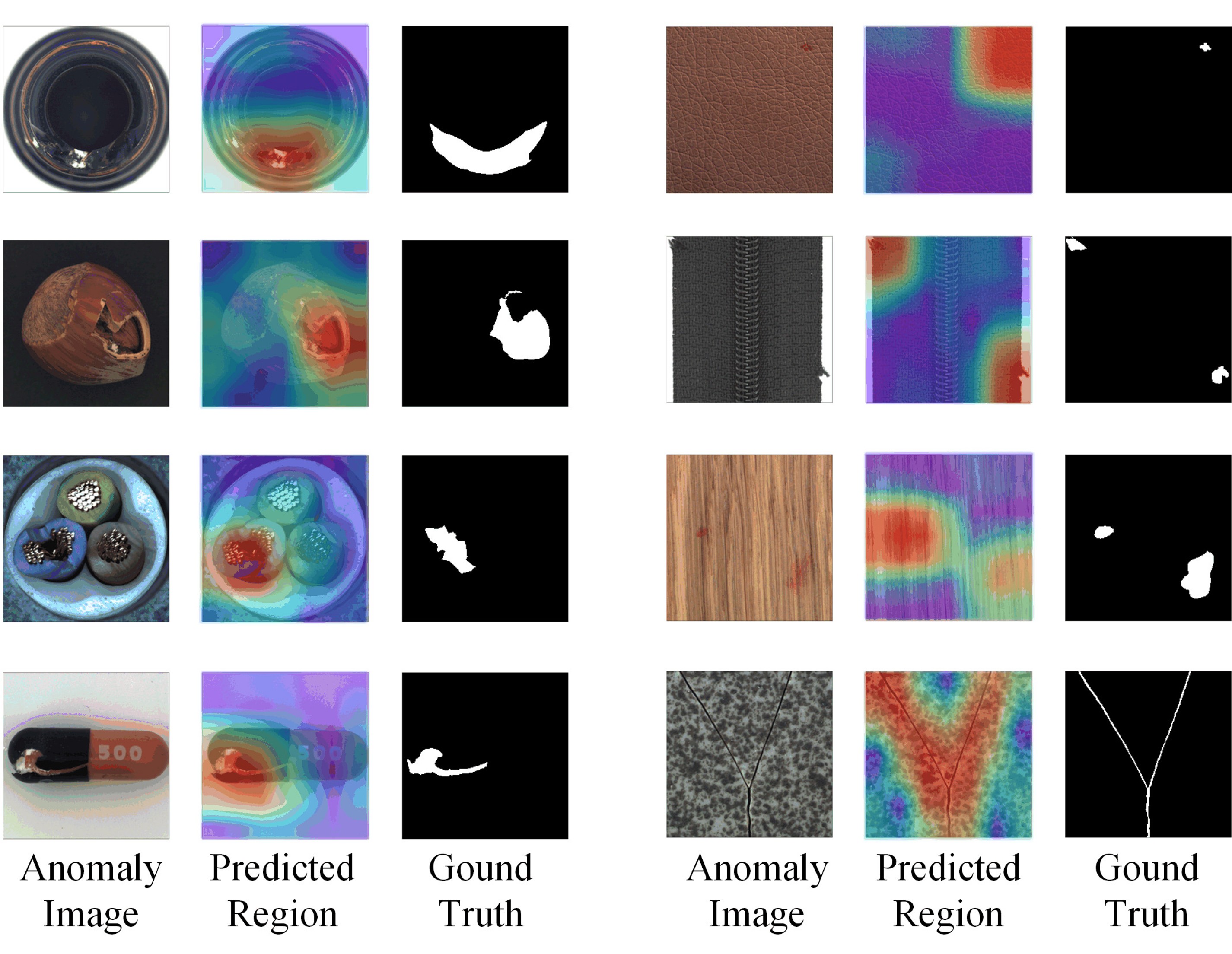}
	\caption{Visualization of MvTec (Sensory AD): With only semantic feature and low resolution feature map, CAP can locate the anomaly area.}
	\label{fig:vis2}
\end{figure}

To show the mechanism of CAP more intuitively, we visualize results referring to class activation mapping(CAM)\cite{zhou2016learning}. In CAM, the product of a global average pooling feature $\mathbf{z}$ and the last layer feature map $\mathbf{f}$ reveals that through which pixels does the model know that the picture belongs to a category. In CAP, $\mathbf{z}_i$ is the global average pooling, thus we compute the element-wise cosine similarity between  $\mathbf{z}_i$ and its corresponding 7*7 pretrained feature map. We also compute the element-wise cosine similarity between $\hat{\mathbf{z}}^n_i$ and the same feature map. These two calculation results are upsampled to 224*224. Naturally, the absolute difference between the two can reflect the anomaly region. We show the visualization results in Figure \ref{fig:vis1} and Figure \ref{fig:vis2}. The dark yellow area represents the anomaly region CAP infers.

\paragraph{Semantic AD}
Semantic AD only focuses on the semantic shift. We show the anomaly region on the normal class and anomaly classes. In the normal class, the anomaly region is some edge pixels and for anomaly classes, the anomaly region is accurately located on the semantic target. This result illustrates that adaptive normal representation obtained by CAP contains the semantic information of the normal class. Meanwhile, this result also shows that the proposed anomaly detection criterion is reasonable and convincing.

\paragraph{Sensory AD}

Sensory AD only focuses on objects with the same or similar semantics and identifies the observational differences on their surface. We select some classes in MvTec and show the calculated anomaly region and the ground truth in Figure \ref{fig:vis2}. The visualization results show that even only the difference between the two semantic features, $\mathbf{z}_i$ and $\hat{\mathbf{z}}^n_i$, is considered, the anomaly region can be located. For anomaly with large region, the localization performance is better than that with smaller region, and the reason for this is that CAP does not leverage decoder mechanism to learn an original size output and the upsampling is through interpolation, thus the inferred anomaly region will be generally larger than authentic region. However, these anomaly regions can all be located more or less with only encodes which implies CAP is an effective and interpretative adaptation strategy for sensory AD task.

\section{Conclusion}

In this paper, we propose a simple yet effective anomaly detection framework. CAP abandons the global optimization goal in traditional one-class setting but focuses on a specific image and its corresponding normal representation. A novel loss function including a simple constraint term is proposed to effectively detect anomalies and avoid pattern collapse. We conduct experiments on multiple benchmarks to verify the effectiveness of CAP, and prove the rationality of each module in the framework through diverse ablation experiments. Adding decoder mechanism to CAP to locate anomaly more precisely is our future research interest.


\clearpage
\bibliographystyle{named}
\bibliography{ijcai22}


\cleardoublepage
\appendix
\noindent{\Large\bfseries Appendix}
\vspace{0.4cm}

\section{Deep-SVDD \& Pattern Collapse}

Deep-SVDD proposes a transformation $G(x;W)$ which maps input data $x \in \mathbb{R}^{D}$ into a new target space where one-class normal data will be compressed into a hypersphere with a predetermined center of \textit{c}, while the anomalies will fall outside the hypersphere. The training objective of one-class Deep-SVDD is defined as
\begin{equation}
	Loss=\frac{1}{N}\sum_{i=1}^{N}\left\|G\left ( x_{i};W \right )-c\right\|^{2}+\frac{\lambda }{2}\sum_{l=1}^{L}\left\| W^{l}\right\|_{F}^{2}
\end{equation}
The first term is a quadratic loss which aims to make the mapping of normal data as close as possible to the spherical center \textit{c}. For a choice of parameters, $W$ is the all-zero network weights, the transformation will map any input to the same output, i.e., $G(x;W)=c_0$ for any $x$. If $c_0$ is equal to \textit{c} in the loss function, Deep-SVDD will fall into a trivial optimal solution which is called hypersphere collapse or pattern collapse.

\section{Expanded Experimental Results}

\subsection{Experiment Configurations}
\paragraph{Dataset Description}
We conduct experiments on various benchmarks: CIFAR-10, CIFAR-100, FMNIST, MvTec. CIFAR-10 and CIFAR-100 are widely used multi-class natural image datasets with 10 and 100 (20 coarse) classes label for each dataset. FMNIST composes of fashion products with 10 categories.  MvTec contains images of 5 unique textures and 10 unique objects from different domains mimicking real-world industrial inspection scenarios. CIFAR-10, CIFAR-100 and FMNIST are semantic anomaly detection. Only one class considered as normal class (like airplane) is accessible in training phase and other classes (like frog, bird, etc.) are seen as anomalies that should be detected in testing phase. MvTec is sensory anomaly detection where defect-free images are intended for training and images with anomalies are intended for testing. All input images are resized to 224 $\times $ 224 pixels to fit pretrained ResNet152 and WideResNet50.

\paragraph{Hyper-parameters Setting}
We fix hyper-parameters referring to the ablation experiments in this paper including $ \mathrm{K} = 32 $, $ \lambda = 2 $, batch size is 64 for CIFAR-10, CIFAR-100 and FMNIST while $ \mathrm{K} = 4 $, $ \lambda = 0.1 $, batch size is 16 for MvTec. We use Adam optimizer with learning rate 5e-4 (1e-4 for MvTec) without decay.

\subsection{Reformed Self-attention Performance}

\begin{figure*}[t]
	\centering
	\includegraphics[scale=.1]{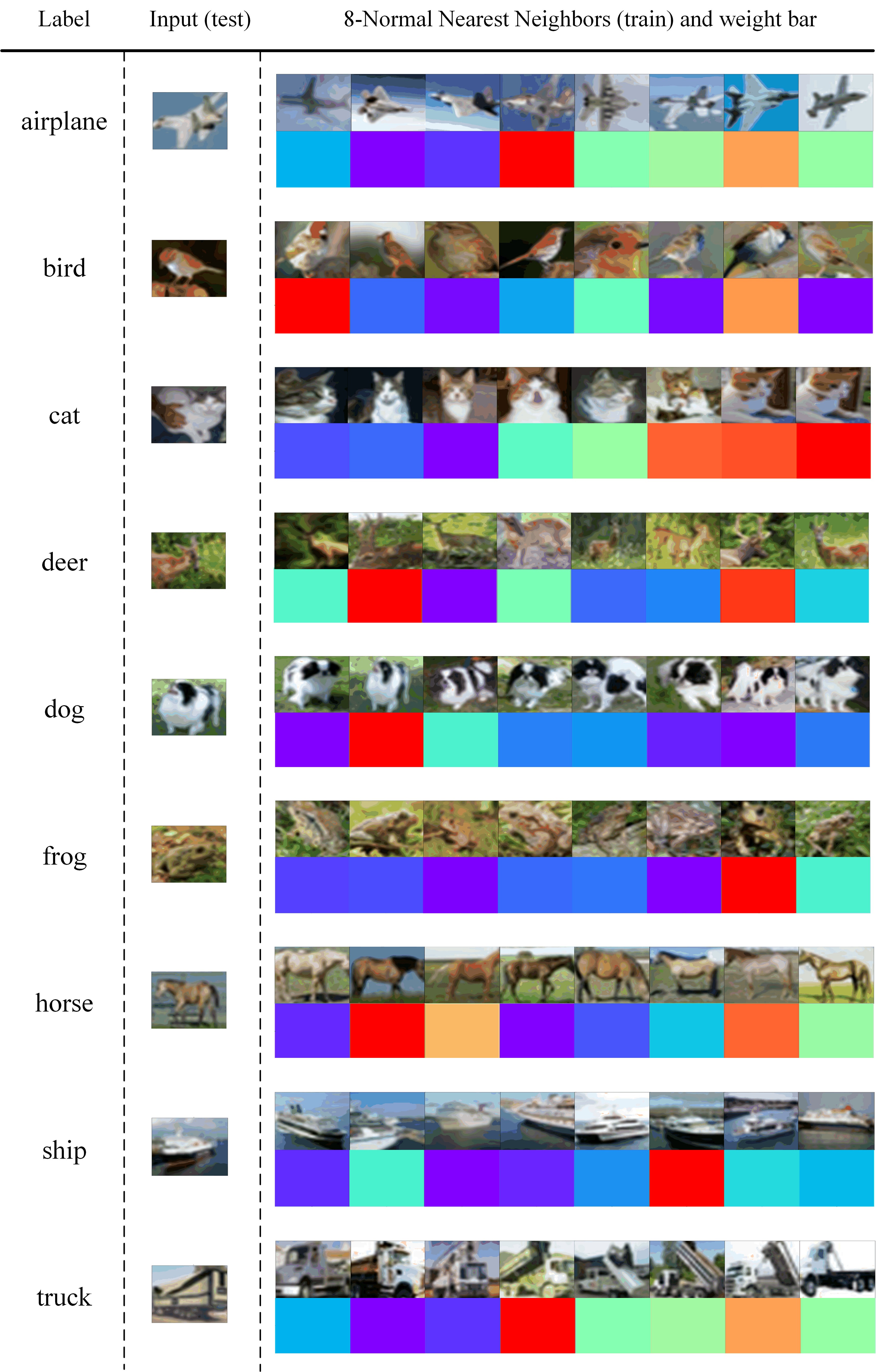}
	\caption{More results on CIFAR-10. The input is sampling from test set and the 8-normal nearest neighbors are traced from memory bank $ \mathbf{M} $. Each row shows the result of selecting a specific class as normal. The input test image is shown on the second column followed by its top-8 most similar normal images of training data and the corresponding weight calculated by reformed self-attention.}
	\label{fig:appendix_1}
\end{figure*}

We visualize the role of reformed self-attention when training on a class (car) in CIFAR-10 in this paper. To show more cases, we train CAP considering the remaining classes in CIFAR-10 as the normal class in turn and show the specific test inputs with their corresponding  8-normal nearest neighbors and weights in Figure \ref{fig:appendix_1}. As the analysis in Section \ref{sec:self-attention}, reformed self-attention provides greater weight for normal nearest neighbors closer to the test input. For these shown images, the neighbor with same object (bird and dog), similar backgrounds (horse, ship) and similar colors (cat, truck) is given the maximum weight. All these visualization results illustrate that the reformed self-attention mechanism provides differentiated weight for $ {\mathbf{M}}_i $ and certain semantic interpretability.

\subsection{Constraint Effect on CIFAR-10 $\&$ CIFAR-100}

\begin{figure}[t]
	\centering
	\includegraphics[scale=.116]{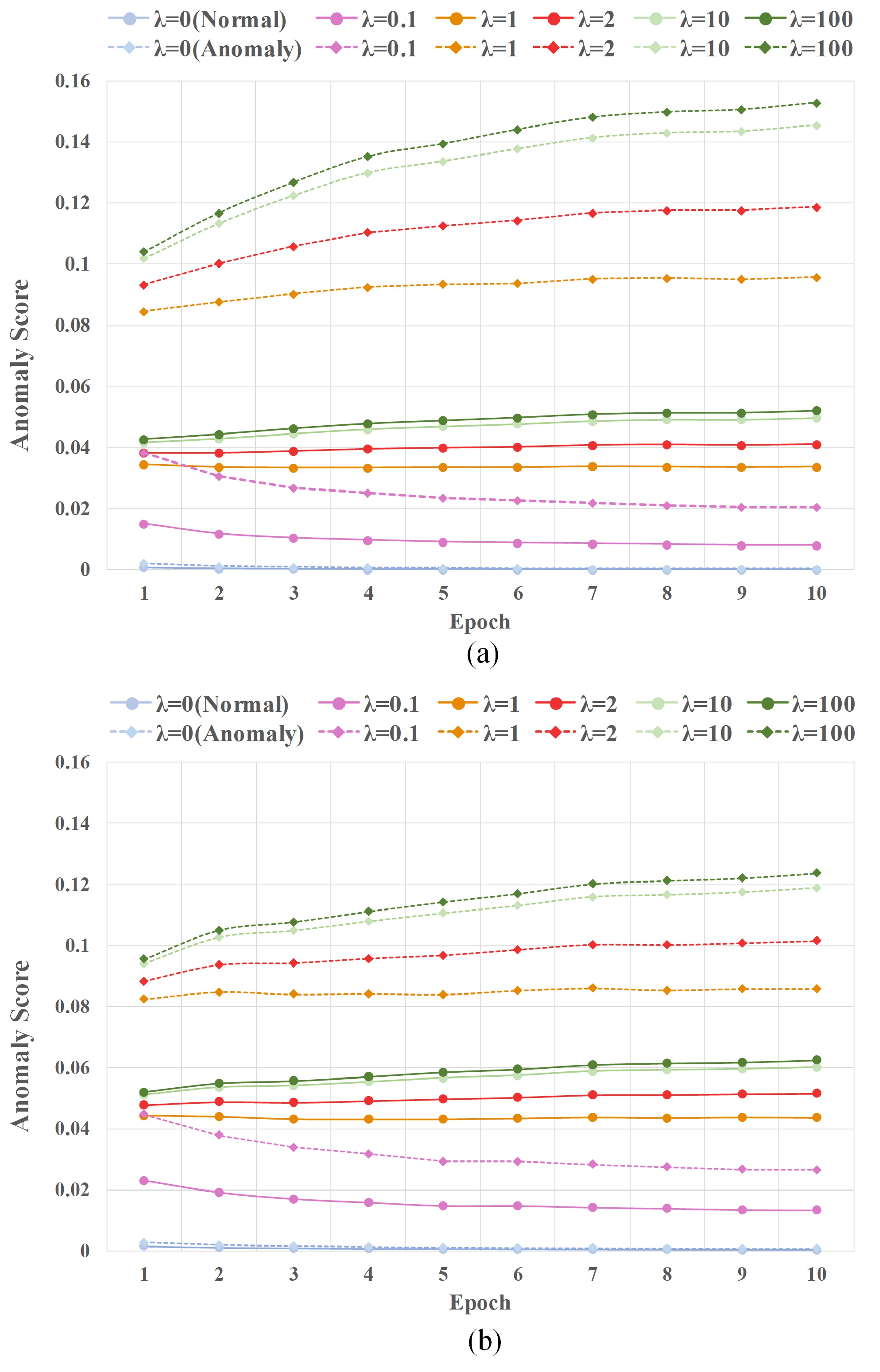}
	\caption{Anomaly score of normal (solid line) and anomaly (dashed line) in test set when training on (\textbf{a}) CIFAR-10 class 1 (bird) and (\textbf{b}) CIFAR-100 class 1 (aquarium\_fish, flatfish, ray, shark and trout) with different $\lambda$.}
	\label{fig:appendix_2}
\end{figure}

To show the effect of constraint term $\Omega$ more persuasively, we show detailed anomaly scores for normal and anomaly when training on CIFAR-10 class 1 (bird) and CIFAR-100 coarse class (aquarium\_fish, flatfish, ray, shark and trout). Figure \ref{fig:appendix_2}.(a) shows basically the same result as that shown in Section \ref{sec:constraint} that small constraint is unable to avoid trivial solution, and large constraint will lead to over-fitting of normal test data. Figure \ref{fig:appendix_2}.(b) shows the results of CAP on CIFAR-100 which reveals the same trend. However, since the label in CIFAR-100 is coarse, the samples in a class are more diverse, so the gap between THE anomaly score of normal and anomaly is small, which is also the reason for the low AUROC of CIFAR-100 compared to CIFAR-10.

\end{document}